\documentclass[10pt,conference,a4paper]{IEEEtran}
\usepackage{times,amsmath,epsfig}
\usepackage{epstopdf}
\title{A Survey of Visual Analysis of Human Motion and Its Applications}
\author{%
	{Qifei Wang}
	\vspace{1.6mm}\\
	\fontsize{10}{10}\selectfont\itshape
	Department of EECS, University of California, Berkeley\\
	University of California, Berkeley, CA 94720, USA\\\\
	\fontsize{9}{9}\selectfont\ttfamily\upshape
	qifei.wang@eecs.berkeley.edu\\
}
\begin{document}
\maketitle

%

\begin{abstract}
This paper summarizes the recent progress in human motion analysis and its applications. In the beginning, we reviewed the motion capture systems and the representation model of human’s motion data. Next, we sketched the advanced human motion data processing technologies, including motion data filtering, temporal alignment, and segmentation. The following parts overview the state-of-the-art approaches of action recognition and dynamics measuring since these two are the most active research areas in human motion analysis. The last part discusses some emerging applications of the human motion analysis in healthcare, human robot interaction, security surveillance, virtual reality and animation. The promising research topics of human motion analysis in the future is also summarized in the last part.
\\[1\baselineskip]
\end{abstract}


\section{Introduction}
\label{sec:intro}
Human motion analysis is one of the most active research areas in artificial intelligence. The body motion is the basis of human activities. The analysis of human motion can help us to learn the sematic information of human activities that includes but not limited to what the subject did which is studied as motion recognition, how the subject performed which is investigated in human performance analysis, and what the subject will do next which is known as motion prediction. These research topics are widely applied in human healthcare, security, and robotics. 

In 2007, Poppe \cite{poppe2007vision} made a comprehensive overview of the vision based human motion analysis. In the past decade, we witnessed significant progress in human motion analysis and its rapid increasing applications. In this paper, we review the most recent updates in human motion analysis, which includes the motion capture, motion representation model, data processing, action recognition, and dynamics measuring. Moreover, we introduce some typical applications of human motion analysis in healthcare, human robot interaction, security surveillance, virtual reality and animation. This overview can help the reader to catch the state-of-the-art technologies and learn the future of human motion analysis. 

The rest of this paper is organized as following: Section \ref{sec:human motion capture and modeling} reviews the motion capture systems and the representation model of human motion data; Section \ref{sec:human motion data processing} sketches the motion data processing, including motion data filtering, temporal alignment, and temporal segmentation; Section \ref{sec:human action recognition} discusses the recent progress in action recognition. Section \ref{sec:learning dynamics} goes through the work of human dynamic measuring and performance analysis; Section \ref{sec:applications} summarizes the emerging applications of human motion analysis and points out the promising research directions in the future.

\section{Human Motion Capture And Modelling}
\label{sec:human motion capture and modeling}
In anthropometry, the human articulated rigid body (HARB) model \cite{kajzer2000human} is widely applied in human motion analysis. Based on the HARB model, the researchers proposed multiple human motion capture approaches and motion representation models. In this section, three popular human motion capture approaches and the representation models are briefly reviewed.

\subsection{Human motion capture}
\label{subsec:human motion capture}
In the early days, human motion is captured by the attached sensors, e.g. accelerometer, gyro, etc \cite{aminian2004capturing}. However, attaching sensor make the data collecting inconvenient and limited in certain applications. The emerging optical sensors introduced some new approaches to record the human motion by visual information \cite{moeslund2006survey}. One basic solution is capturing human motion video by a monocular 2D camera \cite{agarwal2005monocular}. Since humans perform motion in the 3D space, a monocular 2D camera cannot record the motion of every degree-of-freedom (DoF). With the developing of multi-view stereo technologies, multi-camera system \cite{wang2015overview} was brought up to record the human motion in 3D space. One sort of solution uses multiple cameras to track the markers attached on the human’s body and fit the marker coordinates into skeletal joint trajectories \cite{herda2000skeleton}. The marker-based commercial system can provide motion data with sub-millimeter accuracy and frequency of up to 960 Hz \cite{wang2015evaluation}. To avoid the pain of attaching markers, people proposed another sort of markerless multi-view motion capture systems which track the feature points in the video to fit the skeletal joint trajectories \cite{gall2009motion}. Although the multi-view motion capture systems can capture human motion with high accuracy and frame rate, the requirements of capturing stage limit the applications of the multi-view system. 

With the emerging of infrared sensors, the depth camera is introduced to build the affordable human motion capture system \cite{wang2015computational}. The most significant milestone is the release of Microsoft Kinect. The Kinect SDK featured real-time full-body tracking of human limbs based on the algorithm that extract body part from the depth maps \cite{Shotton_2011}. Most recently, Ding and Fan \cite{Meng_HumanPoseEstimation_TIP} proposed a generalized Gaussian kernel correlation function for the similarity measure to facilitate the optimization of body joint angles in a complex articulated structure. The proposed algorithm can also achieve real-time performance with competitive accuracy, which has great potential in mobile applications. 

Compared to the motion capture system, the human pose extracted with Kinect still sufferes from the noise introduced by the depth sensor. In the first version of Kinect, the depth is obtained based on the structured light principle with typical accuracy ranging from about 1-4 cm in the range of 1-4 m. The Kinect version 2 acquires the depth by the time-of-flight principle with improved accuracy that the depth error is under 2 mm in the central viewing cone \cite{Zhang_2012}. Wang et al. \cite{wang2015evaluation} extensively studied the skeletal joint tracking error of Kinect version 1, version 2, and the marker based motion capture system. Based on the analysis, the RGBD camera can obtained relatively stable tracking of human skeletal motion but also introduce jitters on certain joints compared with the high accuracy motion capture system.

\subsection{Modeling}

Human motion is produced by a sophisticated mechanical system composed by bone, muscle, tendon, etc. Fully modelling the human’s motion mechanism is still a puzzle in biomechanics. In the context of computer vision and robotics, a skeletal model is proposed to transform the human motion data into the form of a skeletal joint chain which is represented by the root trajectories and the rotation/translation of each DoF \cite{herda2001using}. The skeletal joint chain model is adaptive to the number of DoF and therefore can be applied to various types of motion. In order to model the mechanism of muscle and tendon, some elastic models are proposed based on the skeletal model \cite{geijtenbeek2013flexible}. The model driven approaches can produce robust analysis of human motion but may lack of smoothness in motion reconstruction due to the constraints on joint DoF. 

Besides the model driven approaches, the model-free approaches which are so-called data driven approaches are also proposed to investigate the human motion. One of the typical applications of the data driven approaches is modeling the skin and muscle deformation in human motion without fitting the motion data into a skeletal model \cite{park2008data}. Compared to the model driven approaches, the data driven approaches can reduce the cost on fitting the motion data to a skeletal model, which sometimes is not efficient to model the human’s pose. However, the data driven approaches suffer from the data dimension burden compared to the parametric model driven approaches.

Detailed comparison of the model driven and data driven approaches is beyond the scope of this paper. In the rest of this paper, the discussion scope is narrowed to the skeletal model approaches.

%

\section{Human Motion Data Processing}
\label{sec:human motion data processing}

\subsection{Motion Data Filtering}
\label{subsec:motion data filtering}
As the introduction in Section II, the accuracy of motion data captured by multi-view camera system is higher than that of the data captured by the RGBD camera, like Microsoft Kinect. The analysis in \cite{wang2015evaluation} reveals that the noise of motion data captured by Microsoft Kinect is due to the loss tracking of human’s body parts. The similar problem exists in the motion data estimated from single 2D camera when the segmentation of human body parts is inaccurate. 

To reduce the outlier in motion data, Wang et al. \cite{wang2015evaluation} proposed a mixture Gaussian and uniform distribution model to fit the joint motion data which classify the regular motion data and the outliers into the Gaussian and uniform distribution, respectively. By applying this mixture model, the variance of the joint position error is reduced by 90\%. 

Other than removing the outlier, Wang et al. \cite{wang2015unsupervised} proposed a kinematic filter based on unscented Kalman filter and the rigid segment length assumption to smooth the jitters in motion data. Applying the proposed filtering to the raw data captured by Kinect significantly reduces the jitters and produces smooth joint trajectories.

The analysis in \cite{wang2015unsupervised} proved that the motion data filtering can help the following data processing and analysis modules produce reliable results based on the noisy input motion data.

\subsection{Temporal Alignment and Dynamic Time Warping}
\label{sec:temporal alignment and dynamic time warping}

Temporal alignment is a challenging problem in the temporal domain signal analysis such as speech recognition, computer vision, and bio-informatics. In human motion analysis, dynamic time warping (DTW) is a widely adopted solution to normalize the paces for the purpose of motion sequence comparison and motion recognition. The classical DTW problem can be solved by the dynamic programming in polynomial time \cite{zhou2012generalized}. Zhou et al. \cite{zhou2012generalized} proposed a generalized time warping to align motion data of different dimensionalities. Kurtek et al. \cite{kurtek2011signal} proposed a random time warping for the purpose of nonlinear signal alignment which shows good results on the repetitive signals like human’s motion.

%
%

\subsection{Temporal Motion Segmentation}
\label{subsec:temproal motion segmentation}
Temporally partitioning continuous motion sequences into atomic actions is another extensively studied topic to facilitate further motion analysis. The general motion segmentation aims at partitioning the motion sequence into multiple primitive atomic actions which constitute the whole motion sequence. To the high-precision motion capture data, Jernej et al. \cite{barbivc2004segmenting} proposed three segmentation methods based on the principle component analysis (PCA) to distinguish one primitive action from the other. Their first two methods can perform the segmentation in real-time using PCA and probabilistic PCA, respectively, whereas the third method (PCA-GMM) fits a Gaussian mixed model to the data of the entire exercise sequence offline. Zhou et al. \cite{zhou2013hierarchical} proposed a bottom-up hierarchical aligned clustering analysis (HACA) algorithm by combining kernel k-means with generalized dynamic time alignment kernel to cluster motion data into motion primitives.  To the noisy motion data captured by the RGBD camera, Sung et al. \cite{sung2012unstructured} proposed a algorithm based on neighbor graph to obtain robust segmentation. 

Another interesting problem is partitioning the repetitive motion sequences into multiple segments where each represents one temporal repetition of the primitive action. Due to the feature within each repetition is almost the same, the general motion segmentation is not practical for this purpose. Ajo et al. \cite{fod2002automated} proposed a zero-velocity crossing (ZVC) detection algorithm based on joint angle velocities to partition the motion data of repetitive arm exercises into individual repetitions. Lu and Ferrier \cite{lu2004repetitive} introduced a multi-dimensional segmentation algorithm to automatically decompose a complex motion into a sequence of simple linear dynamical models. Recently, Wang et al. \cite{wang2015unsupervised} proposed a fully unsupervised algorithm based on the most informative joint selection and adaptive k-means clustering. 

\section{Human Action Recognition}
\label{sec:human action recognition}
Action recognition is one of the most active research areas in computer vision. Based on the input data format, the action recognition approaches can be classified into two classes, video based approaches and skeleton based approaches. Despite of the different input data format, these two types of approaches share the similar framework which usually includes two modules, generating features and training classifier. In this section, we briefly sketch the recent progress of skeleton based motion recognition. More detailed overview can be found in \cite{poppe2010survey}.

\subsection{Feature Generation}
Compared to the video based approaches, the skeleton based approaches can naturally access the skeletal joint trajectories. Extracting the statistical features of joint trajectories followed by a classical classifier, such as nearest neighbour (NN) or support vector machine (SVM) forms the basic solution. Most algorithms in this class use either joint position or joint angles to generate the feature. In \cite{gowayyed2013histogram}, a histogram records the displacements of joint orientations over the whole trajectory. In \cite{ohn2013joint}, a pairwise affinities trajectories of joint angles was proposed to represent the motion. However, these approaches are sensitive to the noise in the motion data. Due to the observation that not all the joint participated in the motion, Ofli et al. \cite{ofli2014sequence} proposed a novel feature representation based on the selection of most informative joints. The experimental results demonstrate that this algorithm can improve the recognition performance by reducing the noise from unrelated joints. This algorithm typically verifies the “less is more” principle in motion recognition. In \cite{vemulapalli2014human}, the human skeletons are transformed to the rotations and translations between different segments which forms a Lie group. Consequently, the human motion is represented as a curved manifold for further classification. This novel representation outperforms most of the existing approaches based on skeletal representation but suffers from the high complexity of transformation. Taking advantage of deep learning, researchers fed the skeletal joint trajectories into the convolutional neural network (CNN) \cite{duskeleton} to train the feature automatically. The features learned from CNN mostly outperform the manually generated feature. To further exploit the temporal features, a hierarchical recurrent neural network (HRNN) \cite{du2015hierarchical} is proposed with high computational efficiency.

\subsection{Probabilistic Graphical Model in Motion Recognition}
\label{subsec:probabilistic}
In the basic motion recognition solution, the classical classifiers do not take advantage of the temporal structural information which makes it not practical for the complex action recognition. The probabilistic graphical model is therefore adopted to exploit the structural feature for human action recognition. The probabilistic graphical models can be divided into two categories: generative models and discriminative models. Hidden Markov model (HMM) is a typical generative model with three assumptions: 1) each action stage is associated with a hidden state and the action is therefore a state transition chain; 2) the current state is only conditioned on the most recent state; 3) the current observation only depends on the current state. HMM uses training data to model the state transition and observation probabilities. In testing, HMM infers the states of the motion sequences and does classification based on the state sequences. Sung et al. \cite{sung2011human} proposed a hierarchical maximum entropy Markov model to the action recognition. However, the independent assumption in HMM which assumes the observations are temporal independent is often not the case.

As opposed to the generative mode, the discriminative model infers the posterior probabilities of the latent action labels given the observations. Therefore, the discriminative model is trained to distinguish the classes rather than learning the parameters in generative models. The discriminative model is more effective than the generative model when the action is similar and the training dataset is large. Conditional random field (CRF) is a typical discriminative model that widely used in human action recognition \cite{wang2009max}. Zhang and Gong \cite{zhang2010action} proposed a hidden CRF (HCRF) which use a single state to label the whole sequences. The HMM pathing introduced in this algorithm can obtain globally optimized parameters of the learned HCRF. Hu et al. \cite{hu2014learning} extended the CRF model with an augmented hidden layer which represents the subtypes of the activities. This additional layer helps to further distinguish the semantic difference such as the action with similar motions but different targets. Experimental results demonstrate this model is efficient in inference than the other existing approach based on CRF.

\section{Learning Dynamics From Motion}
\label{sec:learning dynamics}
Beside learning what people do by action recognition, learning human’s performance in action (e.g. dynamics, stability, flexibility, endurance, etc.) is also an active research area \cite{ofli2016design}. To the best of our knowledge, the non-interventional dynamic measuring is still an open challenge. Since the motion and dynamics is highly intertwined in the physical world, extracting dynamics from motion provides a promising solution to non-interventional dynamic measuring.

Based on the HABR model, Brubaker et al. \cite{brubaker2009estimating} proposed a physical model based on the Newtonian dynamics equation to estimate the contact dynamics. In this model, the dynamics of each rigid body part are modelled by the Newtonian dynamics equation. By applying the principle of virtual work and some assumptions of the contract friction force, they proposed a TMT model associating the joint force and torques with the body mass, inertia, and the motion data. The joint force and torques are therefore solved by the -2 norm optimization. Experimental results verified the estimated torques curves is smooth with consistent standard deviation.  Besides, this approach is extended to track human in motion \cite{brubaker2008kneed}. The results also verify the effectiveness of the dynamics in motion tracking. In \cite{zhang2014leveraging}, Zhang et al. leveraged the wearable pressure sensor to estimate the contact force and extracted the dynamics based on the same principle. They concluded that the dynamics enhanced the realistic visual quality of human motion animation with the object interaction. Agarwal et al. \cite{agarwal2014estimating} also apply the same model to the human motion tracking in video and also achieved robust performance compared to the existing tracking algorithms without dynamics.

\section{Applications and Discussions}
\label{sec:applications}
The applications of human motion analysis are growing fast in the areas including but not limited to healthcare \cite{rao2016anterior}, security, virtual reality \cite{wang2012free}, animation \cite{cao20123d}, human social interaction \cite{ji2014online}, etc., over the past decades. In the first part of this section, we briefly introduce some typical applications of human motion analysis in remote healthcare and human-robot interactions. Next, we point out the future research topics of human motion analysis and its emerging applications. 

In the smart healthcare of elderly and patients with physical problems, monitoring the physical performance remotely can reduce both the cost and risk during the physical training process. In this process, doctors will assign physical training exercises to the subjects and evaluate their performance based on the human motion analysis. To facilitate this process, Ofli et al. \cite{ofli2016design} proposed a remote health coaching system which records the human motion by RGBD camera \cite{wang2010region} and performs online motion performance analysis. The subjects can receive instant feedback from the system followed by the professional feedback from the doctor. Although this kind of systems can provide a solution for remote smart healthcare, there is still a long way from achieving an unsupervised analysis of human motion for the purpose of medicine \cite{wang2015remote}. The motion data accuracy and communication delay \cite{zhai2014content} would be the main burden to make the system adopted by the medical applications.

Another application scenario of human motion analysis is human robot interaction which is desired in both the human daily live and the healthcare. Building robotics to serve human is one of the main objectives of the robotics industry. Understanding human’s language, action, and emotion can improve the intelligence of the robot and make the robot serve people better. Koppula and Saxena \cite{koppula2016anticipating} proposed a human motion prediction algorithm to predict the goal and the trajectories of human’s action. By learning these, the robots can help the human to accomplish the job which is dangerous or arduous. However, due to the high complexity, learning the way that human handle the object is still a challenging problem.

Besides the healthcare and human robot interaction, human motion analysis has also been widely applied to produce high realistic animation of human, enhance the immersion in virtual reality \cite{wang2012complexity} and 3D video \cite{wang2011reduced}\cite{wang2013complexity}, recognize the bad actions to improve the security, control the object in remote surgery. Generally speaking, human motion analysis can be adopted in all the applications that humans participate.

Although extensive efforts have addressed the problems of motion recognition and reconstruction, there are still many unsolved problems in human motion analysis. First of all, capturing motion data with high accuracy by cost-efficient devices is still a challenge problem. Although the existing RGBD camera provided a balance solution, the lack of consistency and low frame rate still requires new solutions coming with the improvements on sensors or algorithms \cite{chen2008wyner}. Human’s motion understanding is another hot research topic beyond the action recognition. The human’s motion contains much high level sematic information including the purpose, emotion, interaction, etc. Extract the sematic information will be super useful in the applications such as human robot interaction, security surveillance, and sociology. Some latest artificial intelligence tools, like CNN would play an important role in this area. Measuring human’s dynamics and its applications in healthcare are still an open research area. Improving the accuracy of dynamic measuring and building a comprehensive model of human physical functionality with respect to the measured dynamics would be most desired features in this area. 

Motion is one of the basic channel for human to interact with the world. Human motion convey various information related to both physiology and psychology. Learning human motion can help to improve the healthcare, social functionality, and security for human. With no doubt, the studies of human motion analysis will raise much attentions and have plenty of applications in the future.

%
%
%

\bibliographystyle{IEEEtran}


\bibliography{biblio}

\begin{thebibliography}{10}
\providecommand{\url}[1]{#1}
\csname url@samestyle\endcsname
\providecommand{\newblock}{\relax}
\providecommand{\bibinfo}[2]{#2}
\providecommand{\BIBentrySTDinterwordspacing}{\spaceskip=0pt\relax}
\providecommand{\BIBentryALTinterwordstretchfactor}{4}
\providecommand{\BIBentryALTinterwordspacing}{\spaceskip=\fontdimen2\font plus
\BIBentryALTinterwordstretchfactor\fontdimen3\font minus
  \fontdimen4\font\relax}
\providecommand{\BIBforeignlanguage}[2]{{%
\expandafter\ifx\csname l@#1\endcsname\relax
\typeout{** WARNING: IEEEtran.bst: No hyphenation pattern has been}%
\typeout{** loaded for the language `#1'. Using the pattern for}%
\typeout{** the default language instead.}%
\else
\language=\csname l@#1\endcsname
\fi
#2}}
\providecommand{\BIBdecl}{\relax}
\BIBdecl

\bibitem{poppe2007vision}
R.~Poppe, ``Vision-based human motion analysis: An overview,'' \emph{Computer
  vision and image understanding}, vol. 108, no.~1, pp. 4--18, 2007.

\bibitem{kajzer2000human}
J.~Kajzer, E.~Tanaka, and H.~Yamada, \emph{Human Biomechanics and Injury
  Prevention}.\hskip 1em plus 0.5em minus 0.4em\relax Springer, 2000.

\bibitem{aminian2004capturing}
K.~Aminian and B.~Najafi, ``Capturing human motion using body-fixed sensors:
  outdoor measurement and clinical applications,'' \emph{Computer animation and
  virtual worlds}, vol.~15, no.~2, pp. 79--94, 2004.

\bibitem{moeslund2006survey}
T.~B. Moeslund, A.~Hilton, and V.~Kr{\"u}ger, ``A survey of advances in
  vision-based human motion capture and analysis,'' \emph{Computer vision and
  image understanding}, vol. 104, no.~2, pp. 90--126, 2006.

\bibitem{agarwal2005monocular}
A.~Agarwal and B.~Triggs, ``Monocular human motion capture with a mixture of
  regressors,'' in \emph{2005 IEEE Computer Society Conference on Computer
  Vision and Pattern Recognition (CVPR'05)-Workshops}.\hskip 1em plus 0.5em
  minus 0.4em\relax IEEE, 2005, pp. 72--72.

\bibitem{wang2015overview}
Q.~Wang, ``An overview of emerging technologies for high efficiency 3d video
  coding,'' \emph{arXiv preprint arXiv:1512.08854}, 2015.

\bibitem{herda2000skeleton}
L.~Herda, P.~Fua, R.~Plankers, R.~Boulic, and D.~Thalmann, ``Skeleton-based
  motion capture for robust reconstruction of human motion,'' in \emph{Computer
  Animation 2000. Proceedings}.\hskip 1em plus 0.5em minus 0.4em\relax IEEE,
  2000, pp. 77--83.

\bibitem{wang2015evaluation}
Q.~Wang, G.~Kurillo, F.~Ofli, and R.~Bajcsy, ``Evaluation of pose tracking
  accuracy in the first and second generations of microsoft kinect,'' in
  \emph{Healthcare Informatics (ICHI), 2015 International Conference on}.\hskip
  1em plus 0.5em minus 0.4em\relax IEEE, 2015, pp. 380--389.

\bibitem{gall2009motion}
J.~Gall, C.~Stoll, E.~De~Aguiar, C.~Theobalt, B.~Rosenhahn, and H.-P. Seidel,
  ``Motion capture using joint skeleton tracking and surface estimation,'' in
  \emph{Computer Vision and Pattern Recognition, 2009. CVPR 2009. IEEE
  Conference on}.\hskip 1em plus 0.5em minus 0.4em\relax IEEE, 2009, pp.
  1746--1753.

\bibitem{wang2015computational}
Q.~Wang, ``Computational models for multiview dense dynamic scene depth maps,''
  \emph{IEEE COMSOC MMTC E-Letter}, vol.~10, no.~5, pp. 16--19, 2015.

\bibitem{Shotton_2011}
J.~Shotton, A.~Fitzgibbon, M.~Cook, T.~Sharp, M.~Finocchio, R.~Moore,
  A.~Kipman, and A.~Blake, ``Real-time human pose recognition in parts from
  single depth images,'' in \emph{Proceedings of the 2011 IEEE Conference on
  Computer Vision and Pattern Recognition (CVPR)}.\hskip 1em plus 0.5em minus
  0.4em\relax Washington, DC, USA: IEEE Computer Society, 2011, pp. 1297--1304.

\bibitem{Meng_HumanPoseEstimation_TIP}
M.~Ding and G.~Fan, ``Articulated and generalized gaussian kernel correlation
  for human pose estimation,'' \emph{IEEE Transactions on Image Processing},
  vol.~25, no.~2, pp. 776--789, Feb 2016.

\bibitem{Zhang_2012}
Z.~Zhang, ``Microsoft {K}inect sensor and its effect,'' \emph{MultiMedia,
  IEEE}, vol.~19, no.~2, pp. 4--10, Feb 2012.

\bibitem{herda2001using}
L.~Herda, P.~Fua, R.~Pl{\"a}nkers, R.~Boulic, and D.~Thalmann, ``Using
  skeleton-based tracking to increase the reliability of optical motion
  capture,'' \emph{Human movement science}, vol.~20, no.~3, pp. 313--341, 2001.

\bibitem{geijtenbeek2013flexible}
T.~Geijtenbeek, M.~van~de Panne, and A.~F. van~der Stappen, ``Flexible
  muscle-based locomotion for bipedal creatures,'' \emph{ACM Transactions on
  Graphics (TOG)}, vol.~32, no.~6, p. 206, 2013.

\bibitem{park2008data}
S.~I. Park and J.~K. Hodgins, ``Data-driven modeling of skin and muscle
  deformation,'' in \emph{ACM Transactions on Graphics (TOG)}, vol.~27,
  no.~3.\hskip 1em plus 0.5em minus 0.4em\relax ACM, 2008, p.~96.

\bibitem{wang2015unsupervised}
Q.~Wang, G.~Kurillo, F.~Ofli, and R.~Bajcsy, ``Unsupervised temporal
  segmentation of repetitive human actions based on kinematic modeling and
  frequency analysis,'' in \emph{3D Vision (3DV), 2015 International Conference
  on}.\hskip 1em plus 0.5em minus 0.4em\relax IEEE, 2015, pp. 562--570.

\bibitem{zhou2012generalized}
F.~Zhou and F.~De~la Torre, ``Generalized time warping for multi-modal
  alignment of human motion,'' in \emph{Computer Vision and Pattern Recognition
  (CVPR), 2012 IEEE Conference on}.\hskip 1em plus 0.5em minus 0.4em\relax
  IEEE, 2012, pp. 1282--1289.

\bibitem{kurtek2011signal}
S.~A. Kurtek, A.~Srivastava, and W.~Wu, ``Signal estimation under random
  time-warpings and nonlinear signal alignment,'' in \emph{Advances in Neural
  Information Processing Systems}, 2011, pp. 675--683.

\bibitem{barbivc2004segmenting}
J.~Barbi{\v{c}}, A.~Safonova, J.-Y. Pan, C.~Faloutsos, J.~K. Hodgins, and N.~S.
  Pollard, ``Segmenting motion capture data into distinct behaviors,'' in
  \emph{Proceedings of Graphics Interface 2004}.\hskip 1em plus 0.5em minus
  0.4em\relax Canadian Human-Computer Communications Society, 2004, pp.
  185--194.

\bibitem{zhou2013hierarchical}
F.~Zhou, F.~De~la Torre, and J.~K. Hodgins, ``Hierarchical aligned cluster
  analysis for temporal clustering of human motion,'' \emph{IEEE Transactions
  on Pattern Analysis and Machine Intelligence}, vol.~35, no.~3, pp. 582--596,
  2013.

\bibitem{sung2012unstructured}
J.~Sung, C.~Ponce, B.~Selman, and A.~Saxena, ``Unstructured human activity
  detection from rgbd images,'' in \emph{Robotics and Automation (ICRA), 2012
  IEEE International Conference on}.\hskip 1em plus 0.5em minus 0.4em\relax
  IEEE, 2012, pp. 842--849.

\bibitem{fod2002automated}
A.~Fod, M.~J. Matari{\'c}, and O.~C. Jenkins, ``Automated derivation of
  primitives for movement classification,'' \emph{Autonomous robots}, vol.~12,
  no.~1, pp. 39--54, 2002.

\bibitem{lu2004repetitive}
C.~Lu and N.~J. Ferrier, ``Repetitive motion analysis: segmentation and event
  classification,'' \emph{IEEE transactions on pattern analysis and machine
  intelligence}, vol.~26, no.~2, pp. 258--263, 2004.

\bibitem{poppe2010survey}
R.~Poppe, ``A survey on vision-based human action recognition,'' \emph{Image
  and vision computing}, vol.~28, no.~6, pp. 976--990, 2010.

\bibitem{gowayyed2013histogram}
M.~A. Gowayyed, M.~Torki, M.~E. Hussein, and M.~El-Saban, ``Histogram of
  oriented displacements (hod): Describing trajectories of human joints for
  action recognition.'' in \emph{IJCAI}, 2013.

\bibitem{ohn2013joint}
E.~Ohn-Bar and M.~Trivedi, ``Joint angles similarities and hog2 for action
  recognition,'' in \emph{Proceedings of the IEEE Conference on Computer Vision
  and Pattern Recognition Workshops}, 2013, pp. 465--470.

\bibitem{ofli2014sequence}
F.~Ofli, R.~Chaudhry, G.~Kurillo, R.~Vidal, and R.~Bajcsy, ``Sequence of the
  most informative joints (smij): A new representation for human skeletal
  action recognition,'' \emph{Journal of Visual Communication and Image
  Representation}, vol.~25, no.~1, pp. 24--38, 2014.

\bibitem{vemulapalli2014human}
R.~Vemulapalli, F.~Arrate, and R.~Chellappa, ``Human action recognition by
  representing 3d skeletons as points in a lie group,'' in \emph{Proceedings of
  the IEEE Conference on Computer Vision and Pattern Recognition}, 2014, pp.
  588--595.

\bibitem{duskeleton}
Y.~Du, Y.~Fu, and L.~Wang, ``Skeleton based action recognition with
  convolutional neural network,'' 2015.

\bibitem{du2015hierarchical}
Y.~Du, W.~Wang, and L.~Wang, ``Hierarchical recurrent neural network for
  skeleton based action recognition,'' in \emph{Proceedings of the IEEE
  Conference on Computer Vision and Pattern Recognition}, 2015, pp. 1110--1118.

\bibitem{sung2011human}
J.~Sung, C.~Ponce, B.~Selman, and A.~Saxena, ``Human activity detection from
  rgbd images.'' \emph{plan, activity, and intent recognition}, vol.~64, 2011.

\bibitem{wang2009max}
Y.~Wang and G.~Mori, ``Max-margin hidden conditional random fields for human
  action recognition,'' in \emph{Computer Vision and Pattern Recognition, 2009.
  CVPR 2009. IEEE Conference on}.\hskip 1em plus 0.5em minus 0.4em\relax IEEE,
  2009, pp. 872--879.

\bibitem{zhang2010action}
J.~Zhang and S.~Gong, ``Action categorization with modified hidden conditional
  random field,'' \emph{Pattern Recognition}, vol.~43, no.~1, pp. 197--203,
  2010.

\bibitem{hu2014learning}
N.~Hu, G.~Englebienne, Z.~Lou, and B.~Kr{\"o}se, ``Learning latent structure
  for activity recognition,'' in \emph{2014 IEEE International Conference on
  Robotics and Automation (ICRA)}.\hskip 1em plus 0.5em minus 0.4em\relax IEEE,
  2014, pp. 1048--1053.

\bibitem{ofli2016design}
F.~Ofli, G.~Kurillo, {\v{S}}.~Obdr{\v{z}}{\'a}lek, R.~Bajcsy, H.~B. Jimison,
  and M.~Pavel, ``Design and evaluation of an interactive exercise coaching
  system for older adults: Lessons learned,'' \emph{IEEE journal of biomedical
  and health informatics}, vol.~20, no.~1, pp. 201--212, 2016.

\bibitem{brubaker2009estimating}
M.~A. Brubaker, L.~Sigal, and D.~J. Fleet, ``Estimating contact dynamics,'' in
  \emph{2009 IEEE 12th International Conference on Computer Vision}.\hskip 1em
  plus 0.5em minus 0.4em\relax IEEE, 2009, pp. 2389--2396.

\bibitem{brubaker2008kneed}
M.~A. Brubaker and D.~J. Fleet, ``The kneed walker for human pose tracking,''
  in \emph{Computer Vision and Pattern Recognition, 2008. CVPR 2008. IEEE
  Conference on}.\hskip 1em plus 0.5em minus 0.4em\relax IEEE, 2008, pp. 1--8.

\bibitem{zhang2014leveraging}
P.~Zhang, K.~Siu, J.~Zhang, C.~K. Liu, and J.~Chai, ``Leveraging depth cameras
  and wearable pressure sensors for full-body kinematics and dynamics
  capture,'' \emph{ACM Transactions on Graphics (TOG)}, vol.~33, no.~6, p. 221,
  2014.

\bibitem{agarwal2014estimating}
P.~Agarwal, S.~Kumar, J.~Ryde, J.~J. Corso, and V.~N. Krovi, ``Estimating
  dynamics on-the-fly using monocular video for vision-based robotics,''
  \emph{IEEE/ASME Transactions on Mechatronics}, vol.~19, no.~4, pp.
  1412--1423, 2014.

\bibitem{rao2016anterior}
Y.~Rao, X.~Ding, J.~Li, J.~Gou, and Q.~Wang, ``Anterior cruciate ligament
  reconstruction model based on anatomical position locating,''
  \emph{Multimedia Tools and Applications}, pp. 1--16, 2016.

\bibitem{wang2012free}
Q.~Wang, X.~Ji, Q.~Dai, and N.~Zhang, ``Free viewpoint video coding with
  rate-distortion analysis,'' \emph{IEEE Transactions on Circuits and Systems
  for Video Technology}, vol.~22, no.~6, pp. 875--889, 2012.

\bibitem{cao20123d}
X.~Cao, Q.~Wang, X.~Ji, and Q.~Dai, ``3d spatial reconstruction and
  communication from vision field,'' in \emph{IEEE International Conference on
  Acoustics, Speech and Signal Processing (ICASSP)}, 2012, pp. 5445--5448.

\bibitem{ji2014online}
X.~Ji, Q.~Wang, B.-W. Chen, S.~Rho, C.~J. Kuo, and Q.~Dai, ``Online
  distribution and interaction of video data in social multimedia network,''
  \emph{Multimedia Tools and Applications}, pp. 1--14, 2014.

\bibitem{wang2010region}
Q.~Wang, X.~Ji, Q.~Dai, and N.~Zhang, ``Region based rate-distortion analysis
  for 3d video coding,'' in \emph{Data Compression Conference (DCC)}, 2010, pp.
  555--555.

\bibitem{wang2015remote}
Q.~Wang, G.~Kurillo, F.~Ofli, and R.~Bajcsy, ``Remote health coaching system
  and human motion data analysis for physical therapy with microsoft kinect,''
  \emph{arXiv preprint arXiv:1512.06492}, 2015.

\bibitem{zhai2014content}
Y.~Zhai, Q.~Wang, Y.~Lu, and S.~Li, ``Content adaptive screen image scaling,''
  in \emph{2014 IEEE International Conference on Image Processing
  (ICIP)}.\hskip 1em plus 0.5em minus 0.4em\relax IEEE, 2014, pp. 3901--3905.

\bibitem{koppula2016anticipating}
H.~S. Koppula and A.~Saxena, ``Anticipating human activities using object
  affordances for reactive robotic response,'' \emph{IEEE transactions on
  pattern analysis and machine intelligence}, vol.~38, no.~1, pp. 14--29, 2016.

\bibitem{wang2012complexity}
Q.~Wang, M.-T. Sun, G.~J. Sullivan, and J.~Li, ``Complexity-reduced geometry
  partition search and high efficiency prediction for video coding,'' in
  \emph{IEEE International Symposium on Circuits and Systems (ISCAS)}, 2012,
  pp. 133--136.

\bibitem{wang2011reduced}
Q.~Wang, J.~Li, G.~J. Sullivan, and M.-T. Sun, ``Reduced-complexity search for
  video coding geometry partitions using texture and depth data,'' in
  \emph{IEEE Visual Communications and Image Processing (VCIP)}, 2011, pp.
  1--4.

\bibitem{wang2013complexity}
Q.~Wang, X.~Ji, M.-T. Sun, G.~J. Sullivan, J.~Li, and Q.~Dai, ``Complexity
  reduction and performance improvement for geometry partitioning in video
  coding,'' \emph{IEEE Transactions on Circuits and Systems for Video
  Technology}, vol.~23, no.~2, pp. 338--352, 2013.

\bibitem{chen2008wyner}
C.~Chen, Q.~Wang, Q.~Dai, Z.~Xiong, and X.~Liu, ``Wyner-ziv coding of 3d
  dynamic meshes,'' in \emph{Electronic Imaging 2008}.\hskip 1em plus 0.5em
  minus 0.4em\relax International Society for Optics and Photonics, 2008, pp.
  68\,221V--68\,221V.

\end{thebibliography}

\end{document}